\documentclass{article}

\usepackage{PRIMEarxiv}

\usepackage[utf8]{inputenc} 
\usepackage[T1]{fontenc}    
\usepackage{hyperref}       
\usepackage{url}            
\usepackage{booktabs}       
\usepackage{amsfonts}       
\usepackage{nicefrac}       
\usepackage{microtype}      
\usepackage{lipsum}

\usepackage{amsmath}
\usepackage{amssymb}
\usepackage{booktabs}
\usepackage{multirow}
\usepackage{diagbox}
\usepackage{xcolor}
\usepackage{graphicx}
\usepackage{subcaption}

\title{Differentiable Search for Finding Optimal Quantization Strategy}

\author{
  Lianqiang Li, Chenqian Yan\\
  ByteDance Inc. \\
  \texttt{\{lilianqiang, yanchenqian.i\}@bytedance.com} \\
   \And
  Yefei Chen \\
  Xi'an Jiaotong University \\
  \texttt{xjtu\_yefeichen@163.com} \\
}

\begin{document}
\maketitle

\begin{abstract}
To accelerate and compress deep neural networks (DNNs), many network quantization algorithms have been proposed. 
Although the quantization strategy of any algorithm from the state-of-the-arts may outperform others in some network architectures, it is hard to prove the strategy is always better than others, and  even cannot judge that the strategy is always the best choice for all layers in a network. In other words, existing quantization algorithms are suboptimal as they ignore the different characteristics of different layers and quantize all layers by a uniform quantization strategy. 
To solve the issue, in this paper, we propose a differentiable quantization strategy search (DQSS) to assign optimal quantization strategy for  individual layer by taking advantages of the benefits of  different quantization algorithms. Specifically, we formulate DQSS as a differentiable neural architecture search problem and adopt an efficient convolution to efficiently explore the mixed quantization strategies from a global perspective by gradient-based optimization. We conduct DQSS for post-training quantization to enable their performance to be comparable with that in full precision models. 
We also employ DQSS in quantization-aware training for further validating the effectiveness of DQSS. To circumvent the expensive  optimization cost when employing DQSS in quantization-aware training, we update the hyper-parameters and the network parameters in a single forward-backward pass. Besides, we adjust the optimization process to avoid the potential under-fitting problem.
Comprehensive experiments on high level computer vision task, i.e., image classification, and low level computer vision task, i.e., image super-resolution, with various network architectures show that DQSS could outperform the state-of-the-arts.
\end{abstract}

\section{Introduction}
\label{sec:intro}
In the past decade, DNNs have gained great success  in many fields, such as computer vision \cite{krizhevsky2012imagenet,he2016deep}, natural language processing \cite{hirschberg2015advances,radford2019language}, and speech processing \cite{graves2013speech,abdel2014convolutional}.
These achievements often rely on the networks with millions or even billions of parameters. With the increasing complexity of network structures, the requirements for storage and computation are becoming considerable challenges. As a result, compressing and accelerating DNNs is imperative, and a wide variety of research efforts have been proposed, including network pruning \cite{li2018filter,lin2019towards}, knowledge distillation \cite{hinton2015distilling,li2021novel}, low-rank approximation \cite{jaderberg2014speeding,tai2015convolutional}, neural architecture search \cite{pham2018efficient,ying2019bench} and network quantization \cite{zhao2021distribution,yamamoto2021learnable}.

Among the above methods, network quantization has attracted great attention \cite{hu2021opq,zhang2021diversifying,boo2021stochastic,zhao2021distribution}, it usually refers to representing the 32-bit floating point (FP32) activations and weights into discrete low-bit integers. Low-precision quantization can not only save the memory storage, but also has the ability to speed up network inference by utilizing inter-arithmetic units if possible. Quantization algorithms can be classified into many categories. According to whether to retain the weights or not, they could be roughly divided as: post-training quantization (PTQ) and quantization-aware training (QAT). However, both the PTQ  and the QAT methods usually  adopt the same quantization strategy for all layers in a network, e.g., all layers employ Kullback-Leibler (KL) divergence to search the quantization threshold. Such uniform quantization strategy assignment may be suboptimal since different layers have different characteristics, and they have different impact on the performance of the overall network. As a result, it is necessary to use mixed quantization strategies for different layers. Nevertheless, few efforts have explored this problem because it requires domain experts with knowledge of machine learning to explore the huge design space smartly with rule-based heuristics \cite{wang2019haq}. Take the ResNet-50 as an example, suppose there are $N=4$ candidate methods for activations and weights in each layer, the size of the search space is about $4^{50}\times 4^{50}$, which is even larger than the number of particles in the earth.

In this paper, we propose a \textbf{d}ifferentiable \textbf{q}uantization \textbf{s}trategy \textbf{s}earch (DQSS) framework to solve this  problem. First, motivated by DARTS \cite{liu2018darts}, DQSS relaxes the discrete search space into a continuous one, so as to explore mixed quantization strategies for different layers from a global perspective by gradient-based optimization. Second, DQSS proposes an efficient convolution rather than perform the operator for each pair of activations and weights like in \cite{liu2018darts}. Theoretically, this enables DQSS to employ all the existing quantization algorithms as the search computation complexity is independent with the size of search space. We integrate DQSS into both the PTQ  and QAT methods. To alleviate the expensive optimization cost brought by QAT, we update the hyper-parameters and the network parameters in a single forward-backward pass. Besides, we improve the optimization process to avoid the under-fitting problem. We conduct extensive experiments on image classification with different neural architectures. We also employ DQSS into image super-resolution to verify its generalizability. Experimental results show that DQSS could surpass previous methods across various network architectures, and even outperforms FP32 models under some circumstances. 

To the best of our knowledge, this is the first work to assign optimal quantization strategy for individual layer (both activations and weights) by taking advantages of the benefits of different quantization algorithms. The contributions of this paper are summarized as following:

\begin{itemize}
\item We propose DQSS to explore mixed quantization strategies for different layers through gradient-based method by formulating assigning optimal quantization strategy as a differentiable neural architecture search problem. 
\item We propose an efficient convolution which enables the search computation complexity of DQSS to be independent with the size of search space.
\item We not only conduct DQSS for the PTQ but also apply it to QAT. Further, we introduce two improvements to circumvent the expensive  optimization cost and avoid the potential under-fitting problem, respectively.
\item We carry out comprehensive experiments and show that the proposed DQSS could outperform the state-of-the-arts for image classification and image super-resolution.
\end{itemize}

The rest of this paper is organized as follows. We summarize some related works in \autoref{relatedwork}. In \autoref{section3}, we first present the motivation of the proposed method, and then introduce the proposed method in detail. We present the experimental results as well as the performance analysis in \autoref{section4}. In \autoref{section5}, we draw the conclusion and describe our future research direction.

\section{Related Works}
\label{relatedwork}
\subsection{Network Quantization}
Network quantization algorithms have long been studied. Take the uniform quantization as an example, a complete quantization process includes quantization operation and de-quantization operation, can be formulated as follows:
\begin{equation}
\label{form1}
T_q=clamp(\lceil \frac{T}{s} \rfloor + z,N_{min}, N_{max})\\ 
\end{equation}
\begin{equation}
\label{form2}
\begin{aligned}
clamp(x,a,b) = a && x \leq a\\
=x && a \leq x \leq b\\
=b && x \geq b
\end{aligned}
\end{equation}
\begin{equation}
\label{form3}
\hat{T} = s * (T_q - z)
\end{equation}
where $T$ is the original FP32 number which has been clipped by quantization threshold. $T_q$ is the quantized number with the range in $[N_{min},N_{max}]$, and $\hat{T}$ is the de-quantized number. $\lceil * \rfloor$ is the rounding function to round continuous numbers to their nearest integers, $s \in \mathbb{R_{+}}$ is called scale which specifies the quantization resolution, and $z \in \mathbb{Z}$ is called zero-point to ensure the common operations like zero padding do not cause quantization error \cite{krishnamoorthi2018quantizing}. Generally, network quantization algorithms could be categorized into post-training quantization and quantization-aware training.

\textbf{Post-training quantization} refers to obtaining quantization parameters, i.e., quantization threshold, scale and zero-point, by a calibration dataset. These methods do not need any fine-tuning or training process and can be deployed easily. Therefore, a lot of works have been explored in both the academia and the industry. Particularly, TensorRT \cite{vanholder2016efficient} employed maximum absolute value method (Max\_Abs) and KL divergence between the FP32 numbers and the quantized numbers to gain the quantization parameters.
\cite{wu2020easyquant} proposed EasyQuant (EQ), a method tried to maximize the cosine similarity between the FP32 numbers and the quantized numbers.
\cite{nagel2019data} introduced a data-free quantization method by equalizing the weights range and correcting the bias errors that were introduced during quantization. \cite{nagel2020up} proposed AdaRound, a better weight-rounding mechanism for PTQ that adapts to the data and the task loss. 
\cite{leng2018extremely} borrowed the idea from alternating direction method of multipliers (ADMM) to decouple the optimization of minimizing quantization error into several sub-problems.
In spite of so many works have explored, the performance of PTQ is usually worse than that in QAT.

\textbf{Quantization-aware Training} involves optimizing the weights to adapt quantization noise by using a training set. Rendering the training ``quantization-aware" is non-trivial since the rounding and clipping operations in Eq. (\ref{form1}) are non-differentiable \cite{finkelstein2019fighting}. The straight-through estimator (STE) \cite{bengio2013estimating} trick can be used to approximate the gradient calculation for backpropagation. Previous works, e.g., DoReFa \cite{zhou2016dorefa} quantized activations and weights in a deterministic way while quantized the gradients stochastically. Recently, more and more works tried to learn the quantization information during the 
fine-tuning process for quantized networks. To solve the gradients mismatching problem, DSQ \cite{gong2019differentiable} proposed a differentiable soft quantization method to provide more accurate gradients compared with STE. LSQ \cite{esser2019learned} introduced a novel approach to learn quantization resolution and network parameters conjunctively. PACT \cite{choi2018pact} proposed to adaptively learn and determine the quantization threshold during the fine-tuning process. 
QAT methods can provide higher performance than that in PTQ, but the fine-tuning process is usually time-consuming and computationally intensive \cite{zhang2021diversifying}. 

Compared with the proposed DQSS, both of the existing PTQ methods and QAT methods simply assign the same quantization strategy to all layers, which usually leads to the suboptimal solution.

\subsection{Neural Architecture Search} Neural architecture search is also a mainstream approach for lightweight DNNs. The first work \cite{zoph2016neural} used an reinforcement learning agent to generate neural network architectures with high accuracy. However, the method required thousands of GPU days. To efficiently search for neural architecture, an attractive solution was to construct a super network whose weights are shared with its child networks \cite{pham2018efficient,bender2018understanding}. These methods  could sample relatively better child networks from the super network by another network controller. Recently, DARTS shortened \cite{liu2018darts} the searching time to a few GPU days by relaxing the discrete architecture search space to be continuous with weighted mixture-of-operations, and optimized the candidate architecture through gradient descent method. 

The neural architecture search algorithms can also benefit other tasks. Similar to our work, the combination of network quantization and neural architecture search has been actively studied in recent literatures\cite{yu2020search,li2020efficient,cai2020rethinking,wang2020differentiable}. However, they pay more attention to the mixed precisions for different layers while DQSS manages to find the mixed quantization strategies for different layers.

\section{The Proposed Method}
\label{section3}

In this section, we first present the motivation of DQSS. Then we introduce DQSS in detail, including how to formulate the problem of searching mixed quantization strategies by a differentiable way, how to carry out efficient convolution, and how to solve the problems brought by employing DQSS in QAT.

\subsection{Motivation}
The core of quantization strategy is to find a proper quantization threshold to set up scale and zero-point. However, there exists a trade-off between the threshold and the scale value. Specifically, to increase the quantization resolution, the threshold should be as small as possible. However, if the quantization threshold is too small, it may remove valid values outside the threshold which would degrade the performance of quantized networks.

\begin{figure}[!htb]
\centering
\begin{subfigure}{0.48\textwidth}
\centering
\includegraphics[width=\textwidth]{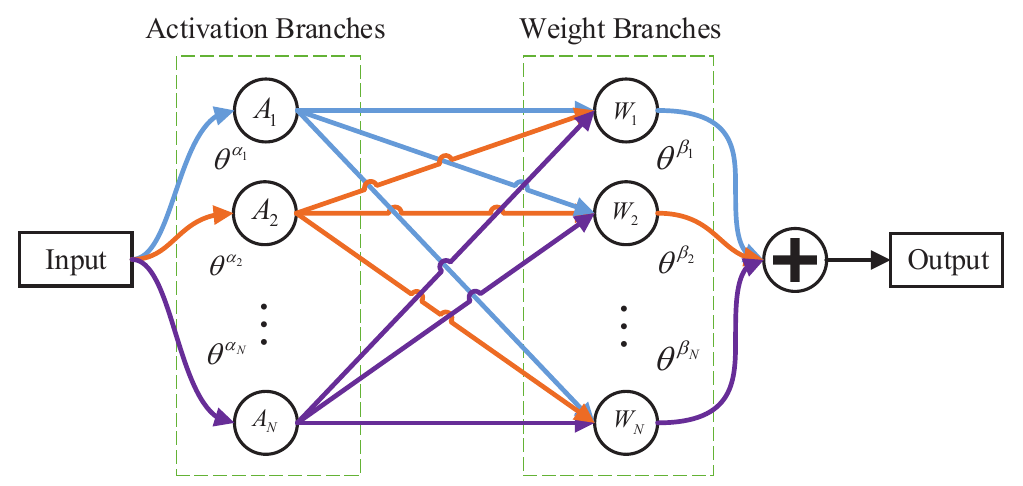}
\caption{The computation graph similar to the naive DARTS}
\label{f11}
\end{subfigure}
\begin{subfigure}{0.48\textwidth}
\centering
\includegraphics[width=\textwidth]{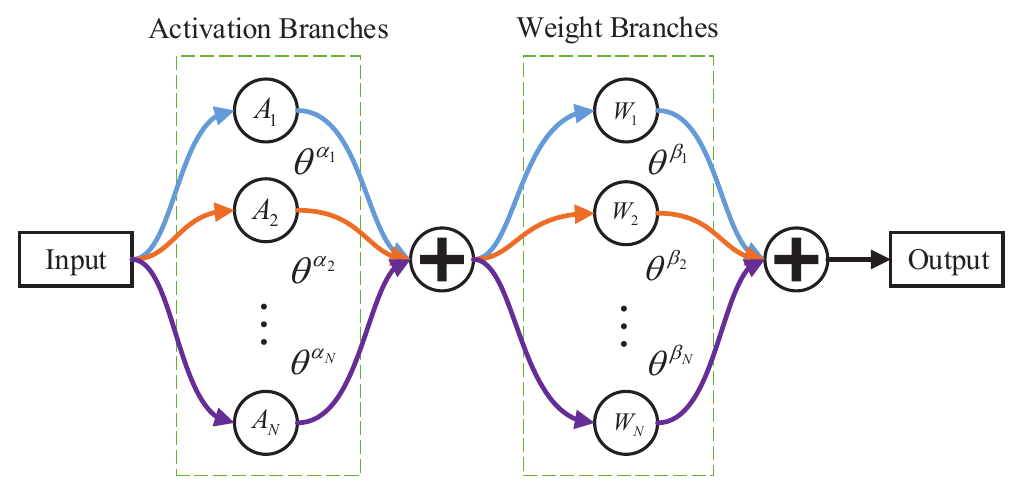}
\caption{The computation graph of the proposed DQSS}
\label{f12}
\end{subfigure}
\caption{The illustrations of different computation graphs. For the computation graph similar to the naive DARTS, each quantized activation branch and each quantized weight branch need carry on a convolution. For DQSS, there is only one convolution between the weighted average sum of all quantized activation branches and the weighted average sum of all quantized weight branches.}
\label{f1}
\end{figure}

Although most of the state-of-the-arts \cite{zhou2016dorefa,choi2018pact,esser2019learned,gong2019differentiable} claimed they could balance the above trade-off and gain satisfactory performance. Some researchers \cite{li2021mqbench} recently found that  no algorithm could achieve the best performance on every architecture. They also noticed that employing different quantization strategies for activations and weights could gain better performance than employing a uniform one. Echoing the findings of academia, in industry, some neural network inference frameworks, e.g. TensorRT \cite{vanholder2016efficient}, would expose some interfaces to users. By employing the interfaces to modify the quantization information manually, users could alleviate the performance drop caused by quantization. Nonetheless, as DNNs become deeper, the search space increases exponentially, which makes it infeasible to rely on hand-crafted schemes. 

We attribute these phenomena to that the existing quantization algorithms ignore the different characteristics of different layers and quantize all layers by a uniform quantization strategy. 
Motivated by the above observations and the related works, in this paper, we look forward to finding an optimal quantization strategy for each individual layer by taking advantages of the benefits of different quantization algorithms.

\subsection{Differentiable quantization strategy search}
Without loss of generality, we take the convolution operator as a concrete example in the following description. Denote $A$ as the activations (inputs) while $W$ as the weights. Then the conventional convolution can be calculated by follows:
\begin{equation} 
\label{form4}
y=W * A\\ 
\end{equation}
where $y$ are the outputs. Different with the conventional convolution, for each layer, we construct $N$ quantized branches for the activations and $N$ quantized branches for the weights. Where $N$ denotes the number of candidate quantization strategies. Then the optimization objective of DQSS is to search the optimal quantization strategy for each layer from the given candidate methods to keep the performance of quantized models. A straightforward solution is to learn an importance parameter for each branch, this can be implemented by reformulating Eq. (\ref{form4}) as:
\begin{equation} 
\begin{aligned}
\label{form5}
y~& = \sum_{i=1}^{N}{\alpha }_{i} \sum_{j=1}^{N}{\beta}_{j} (W_{j}*A_{i}),  \\
&s.t. \sum {\alpha }_{i} = 1,\sum {\beta }_{j} = 1
\end{aligned}
\end{equation}
where $\alpha$ and $\beta$ are the importance parameters for activations and weights. 
As different branches can have different impact on the overall performance, we can train the importance parameters to increase the probability to sample those branches with better performance, and to suppress those with worse performance.
If the networks with multi-branches have been converged, then we can take the "winner-take-all" scheme to derive the optimal quantization strategy assignment as:
\begin{equation}
\label{form6}
\begin{aligned}
\alpha_{*} = argmax(\alpha),~~\beta_{*} = argmax(\beta)
\end{aligned}
\end{equation}

However, since the $argmax$ function is not differentiable and the search space is discrete and huge, we cannot optimize the importance parameters by hand-craft approaches or using gradient-based methods. Inspired by DARTS \cite{liu2018darts}, a much simpler optimization is to relax the discrete search space into a continuous one by applying the $Softmax$ function over the learnable importance parameters. As a result, we can rewrite the Eq.~(\ref{form5}) as follows:
\begin{equation}
\begin{aligned}
\label{form7}
y& = \sum_{i=1}^{N}\theta^{\alpha_{i}} \sum_{j=1}^{N}\theta^{\beta_{j}} (W_{j}*A_{i}),  \\
&s.t. \sum \theta^{\alpha_{i}} = 1,\sum \theta^{\beta_{j}} = 1
\end{aligned}
\end{equation}
where $\theta^{\alpha}$ and $\theta^{\beta}$ are the softened importance parameters, and their definitions are shown in Eq.~(\ref{form8})
\begin{equation}
\label{form8}
\theta^{\alpha_{i}} = \frac{exp(\alpha_{i})}{\sum_{k}(\alpha_{k})},~~
\theta^{\beta_{j}} = \frac{exp(\beta_{j})}{\sum_{k}(\beta_{k})}
\end{equation}

Such relaxation enables the optimization process can be learned by gradient descent method in the space of continuous parameters $ \{ \theta^{\alpha} , \theta^{\beta} \}$, which is much less expensive than combinatorial search over the configurations of $\{{\alpha},{\beta}\}$.

\subsection{Efficient convolution}
Although naive DARTS enables to search mixed quantization strategies to be feasible, we have to store $N$ copies of activations and $N$ copies of weights to perform Eq.~(\ref{form7}), as shown in \autoref{f11}. Besides the storage requirement, the computation cost in Eq.~(\ref{form7}) is as much as the $N\times N$ times than that in Eq.~(\ref{form4}). Fortunately, unlike naive DARTS, where the candidate operators are heterogeneous, e.g. convolution, skip connection, pooling, etc., the candidate operators of DQSS are homogeneous, only difference is that different branches would carry out different quantization strategies. Thus, we only need keep the representative activation branch and the representative weight branch as illustrated in \autoref{f12}, to avoid the expensive parallel operations. Specifically, the quantized activation tensors of different quantization strategies are scaled and then summed as $\hat{\textbf{A}}$ and the quantized weight tensors of different quantization strategies are scaled and then summed as $\hat{\textbf{W}}$, their definitions are shown in Eq. (\ref{form9})
\begin{equation}
\label{form9}
\hat{\textbf{A}}= \sum^{N}_{i=1}\theta^{{\alpha}_{i}}A_{i},~\hat{\textbf{W}} = \sum^{N}_{j=1}\theta^{{\beta}_{j}}W_j
\end{equation}

Therefore, we can reformulate Eq.~(\ref{form7}) as follows:
\begin{equation}
\begin{aligned}
\label{form10}
y = (\sum^{N}_{j=1} \theta &^{{\beta}_{j}}W_j)* (\sum^{N}_{i=1}\theta^{{\alpha}_{i}}A_{i} )=\hat{\textbf{W}} * \hat{\textbf{A}},  \\
s.&t. \sum \theta^{\alpha_{i}} = 1,\sum \theta^{\beta_{j}} = 1
\end{aligned}
\end{equation}

Compared with Eq.~(\ref{form7}), Eq.~(\ref{form10}) improves the memory cost from $O(N)$ to $O(1)$, and the computational cost from $O(N^2)$ to $O(1)$. Theoretically, this enables DQSS to be able to employ all the existing quantization algorithms to find the optimal quantization strategy assignment as the search computation complexity of DQSS is independent with the size of search space.

In this paper, we mainly focus on applying DQSS to PTQ methods. This means that the optimization process just need learn the hyper-parameters, i.e., $\theta^{\alpha}$ and $\theta^{\beta}$, which is relatively easier than that in QAT methods.

\subsection{Extension to quantization-aware training}
For further validating the effectiveness of DQSS, we also perform it in QAT methods. Nevertheless, there exist two issues. On the one hand, as the QAT methods need fine-tuning process, both the hyper-parameters and network parameters need be learned by gradient descent. For neural architecture search, this problem can be solved by a second-order bi-level optimization \cite{anandalingam1992hierarchical}. However, such solution is not suitable for DQSS and this can be time consuming when searching the optimal quantization strategy for large models. On the other hand, when carrying on the backward pass for Eq.~(\ref{form10}), the gradients for each layer will be distributed to all candidate branches. This leads to the branch with low importance parameter $\theta^{\beta_{j}}$ would only receive a few gradients, and the network may be under-trained.

To avoid the above issues, we propose two corresponding solutions. For the bi-level optimization problem, we treat hyper-parameters and network parameters equally, so as to update them in a single forward-backward pass. For the second issue, suggested by \cite{cai2020rethinking}, we let all branches share all the weights by replacing $W_i$ with ${\textbf{W}}$, and redefine $\hat{\textbf{W}}$ as:
\begin{equation}
\label{form11}
\hat{\textbf{W}} = \sum^{N}_{j=1}\theta^{\beta_{j}}\textbf{W}
\end{equation}
where $\textbf{W}$ is the universal weight tensor.  Although the gradients may be still distributed to each branch, they are all accumulated to update $\textbf{W}$ . This can eliminate the potential under-fitting problem \cite{cai2020rethinking}.

\section{Experiments}
\label{section4}
In this section, we firstly evaluate our proposed DQSS on PTQ. Then we apply DQSS to QAT when networks are quantized by ultra-low bitwidths. At last, we present some ablation studies. To make a fair comparison, all experiments are conducted on PyTorch framework \cite{paszke2019pytorch}. We apply the same quantization settings, i.e., per-tensor, symmetric and without any BatchNorm folding, to all experiments. The only difference is that 8-bit quantization is for PTQ while 4-bit quantization is for QAT.

\begin{table*}[htbp]
\small
\centering
\caption{\label{t1} The comparison of the Top-1 accuracy performance (\%) for the 8-bit PTQ experiments on the  ImageNet classification. In each column, \textcolor{red}{red} / \textcolor{blue}{blue} represents \textcolor{red}{best} / \textcolor{blue}{second best}, respectively. }
\begin{tabular}{ccccccc}
\hline
\multirow {2} {*} {\diagbox{Methods}{Models}} & \multirow {2} {*} {VGG-16} & \multirow {2} {*} {ResNet-18} & \multirow {2} {*} {ResNet-50} & \multirow {2} {*} {ShuffleNet-V2} & \multirow {2} {*} {MobileNet-v2} & \multirow {2} {*} {Average architecture}  \\
\\
\hline
FP32 Baseline & 71.592 & 69.758 & 76.130 & 69.362 & 71.878  & 71.744   \\
Max\_Abs\cite{vanholder2016efficient} & 71.434 & 69.198 & 74.912 & 64.290 & 63.440 &68.655  \\
KL\cite{vanholder2016efficient}& \textcolor{blue}{71.536} & 69.430 & \textcolor{blue}{75.473} & 64.718 & 67.604 & 69.752 \\
EQ\cite{wu2020easyquant}& 71.338 & 69.391 & 75.182 & 64.158 & 66.838 & 69.381  \\
ADMM \cite{leng2018extremely} & 71.496 & \textcolor{blue}{69.497} & 75.406 & \textcolor{blue}{66.836} & \textcolor{blue}{70.413} & \textcolor{blue}{70.730} \\
DQSS & \textcolor{red}{71.842} & \textcolor{red}{70.294} & \textcolor{red}{75.679} & \textcolor{red}{68.442} & \textcolor{red}{71.628} & \textcolor{red}{71.577} \\
\hline
\end{tabular}
\end{table*}
\begin{table*}
\small
\centering
\caption{Quantitative comparison (average PSNR/SSIM) with the state-of-the-art quantization methods for image super-resolution of 4x upscaling on the public benchmark. MSRResNet is the backbone network. In each column, \textcolor{red}{red} / \textcolor{blue}{blue} represents \textcolor{red}{best} / \textcolor{blue}{second best}, respectively.}
\label{tab:sr_MSRResNet_mx4}
\begin{tabular}{ccccccc}
\toprule
\multirow {2} {*} {\diagbox{Methods}{Datasets}} & \multirow {2} {*} {Set5} & \multirow {2} {*} {Set14} & \multirow {2} {*} {BSD100} & \multirow {2} {*} {Urban100} & \multirow {2} {*} {Manga109}   & \multirow {2} {*} {Average dataset } \\
\\
\midrule
FP32  Baseline    &  30.25/0.8651 &  26.78/0.7451 &  26.23/0.7116 &  24.60/0.7658 &  28.54/0.8807 &  27.28/0.79366 \\
Max\_Abs\cite{vanholder2016efficient}       &  30.14/0.8613 &  26.68/0.7415 &  26.16/0.7086 &  \textcolor{red}{24.53}/0.7619 &  \textcolor{blue}{28.34/0.8761} &  27.17/0.78988 \\
KL\cite{vanholder2016efficient}        &  30.13/0.8621 &  26.67/0.7419 &  26.17/0.7091 &  24.47/0.7613 &  28.26/0.8758 &  27.14/0.79004 \\
EQ\cite{wu2020easyquant}       &  \textcolor{blue}{30.15/0.8627} &  \textcolor{blue}{26.70/0.7423} &  \textcolor{blue}{26.18/0.7092} &  24.52/\textcolor{blue}{0.7624} &  28.31/0.8767 &  \textcolor{blue}{27.17/0.79066} \\
ADMM\cite{leng2018extremely}      &  29.95/0.8605 &  26.56/0.7396 &  26.11/0.7076 &  24.29/0.7570 &  27.78/0.8706 &  26.94/0.78706 \\
DQSS &  \textcolor{red}{30.18/0.8628} &  \textcolor{red}{26.72/0.7426} &  \textcolor{red}{26.18/0.7093} &  \textcolor{blue}{24.52}/\textcolor{red}{0.7625} &  \textcolor{red}{28.39/0.8775} &  \textcolor{red}{27.20/0.79094} \\
\bottomrule
\end{tabular}
\end{table*}
\begin{table*}
\small
\centering
\caption{Quantitative comparison (average PSNR/SSIM) with the state-of-the-art quantization methods for image super-resolution of 4x upscaling on the public benchmark. EDSR is the backbone network. In each column, \textcolor{red}{red} / \textcolor{blue}{blue} represents \textcolor{red}{best} / \textcolor{blue}{second best}, respectively.}
\label{tab:sr_EDSR_mx4}
\begin{tabular}{ccccccc}
\toprule
\multirow {2} {*} {\diagbox{Methods}{Datasets}} & \multirow {2} {*} {Set5} & \multirow {2} {*} {Set14} & \multirow {2} {*} {BSD100} & \multirow {2} {*} {Urban100} & \multirow {2} {*} {Manga109}   & \multirow {2} {*} {Average dataset } \\
\\
\midrule
FP32 Baseline   &  30.17/0.8641 &  26.72/0.7430 &  26.22/0.7109 &  24.50/0.7619 &  28.48/0.8779 &  27.22/0.79156 \\
Max\_Abs\cite{vanholder2016efficient}    &  30.01/0.8544 &  26.63/0.7356 &  26.13/0.7035 &  \textcolor{blue}{24.42}/0.7532 &  \textcolor{red}{28.32}/0.8679 &  27.10/0.78292 \\
KL\cite{vanholder2016efficient}     &  30.08/\textcolor{blue}{0.8609} &  26.62/0.7396 &  \textcolor{blue}{26.17/0.7086} &  24.33/0.7564 &  27.73/0.8667 &  26.99/0.78644 \\
EQ\cite{wu2020easyquant}    &  30.08/0.8589 &  26.65/0.7387 &  26.16/0.7066 &  24.41/0.7562 &  \textcolor{blue}{28.31}/0.8714 &  27.12/0.78636 \\
ADMM\cite{leng2018extremely}   &  \textcolor{blue}{30.10}/0.8606 &  \textcolor{blue}{26.65/0.7400} &  26.17/0.7083 &  24.41/\textcolor{blue}{0.7574} &  28.25/\textcolor{blue}{0.8726} &  \textcolor{blue}{27.12/0.78778} \\
DQSS    &  \textcolor{red}{30.12/0.8614} &  \textcolor{red}{26.67/0.7408} &  \textcolor{red}{26.18/0.7090} &  \textcolor{red}{24.45/0.7591} &  28.30/\textcolor{red}{0.8738} &  \textcolor{red}{27.14/0.78882} \\
\bottomrule
\end{tabular}
\end{table*}

\subsection{Experiments on PTQ}
\label{section41}
In this subsection, we not only evaluate DQSS on image classification but also carry out experiments on image super-resolution.
As for the candidate methods for DQSS, we choose  Max\_Abs\cite{vanholder2016efficient}, KL\cite{vanholder2016efficient}, EQ\cite{wu2020easyquant} and ADMM\cite{leng2018extremely} to set up the candidate pool.
Both the $\alpha_i$ and $\beta_j$ are initialized as 0.1, thus the $\theta^{\alpha_i}$ and the $\theta^{\beta_j}$ are initialized as 0.25 according to Eq.~(\ref{form8}) as there are four methods. The calibration dataset includes 256 images which are randomly selected from the corresponding task's training set. 
We use SGD optimizer for learning the hyper-parameters $\theta$. 
The learning rate $lr_{\theta}$ is initialized to be $10^{-4}$.
We train a maximum 3 epochs and decay the learning rate by 10 at the beginning of the second and the third epochs.

\textbf{Experiments on image classification:}
ImageNet classification task \cite{deng2009imagenet} is the widely used benchmark in image classification. We conduct experiments using various networks including VGG-16, ResNet-18, ResNet-50, ShuffleNet-V2 and MobileNet-V2.  The pretrained models are downloaded from the official model zoo from torchvision \cite{paszke2019pytorch}. 
The training images are resized, cropped to $224 \times 224$ pixels and randomly flipped horizontally with the batch size of 256. The testing images are center-cropped to $224 \times 224 $ pixels. 

\begin{figure*}[!htb] 
   \centering
   \begin{subfigure}[H]{0.13\linewidth} 
    \subcaption*{\centerline{GT}} 
    \includegraphics[width=0.95\linewidth, height=1.3\linewidth]{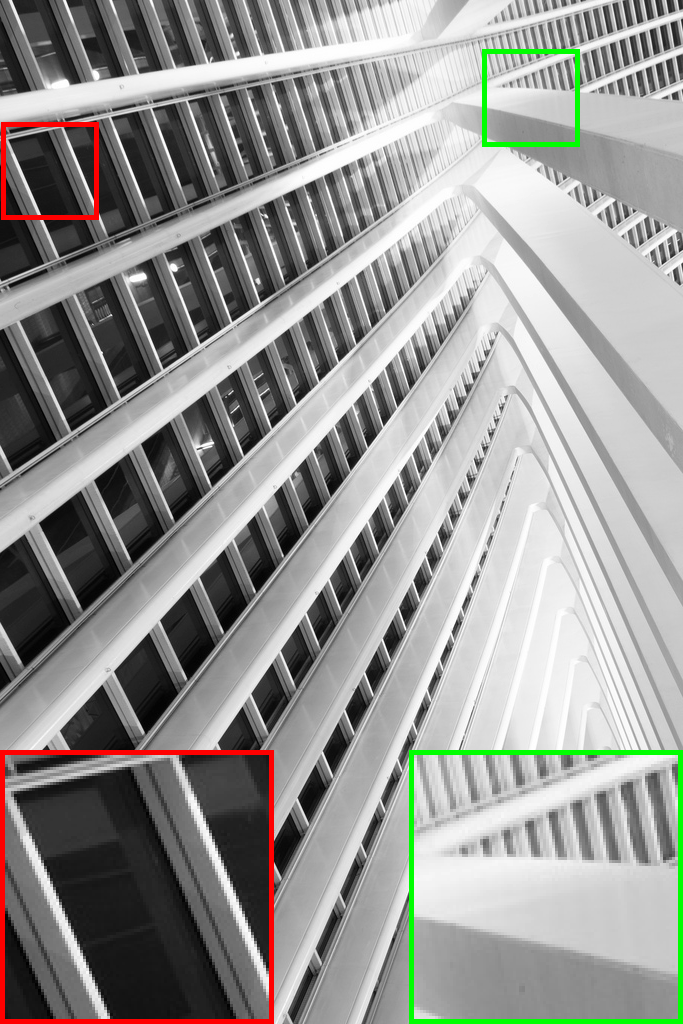} \\ 
    \subcaption*{\centerline{PSNR / SSIM}} 
   \end{subfigure}
   \begin{subfigure}[H]{0.13\linewidth} 
    \subcaption*{\centerline{FP32}} 
    \includegraphics[width=0.95\linewidth, height=1.3\linewidth]{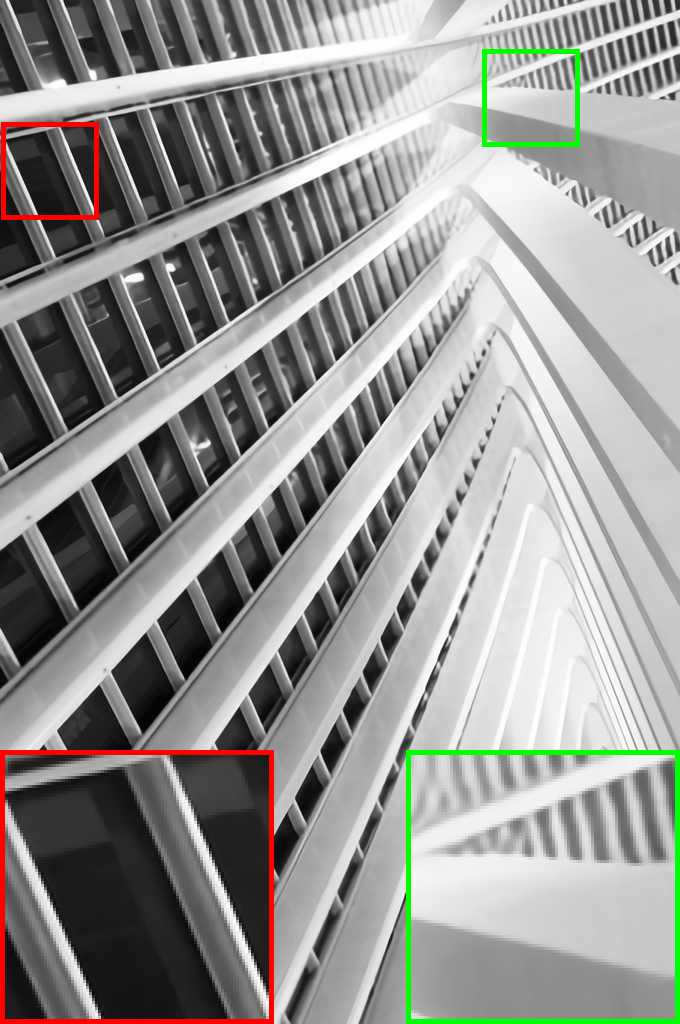} \\ 
    \subcaption*{\centerline{27.79/0.8829}} 
   \end{subfigure}
   \begin{subfigure}[H]{0.13\linewidth} 
    \subcaption*{\centerline{Max\_Abs}} 
    \includegraphics[width=0.95\linewidth, height=1.3\linewidth]{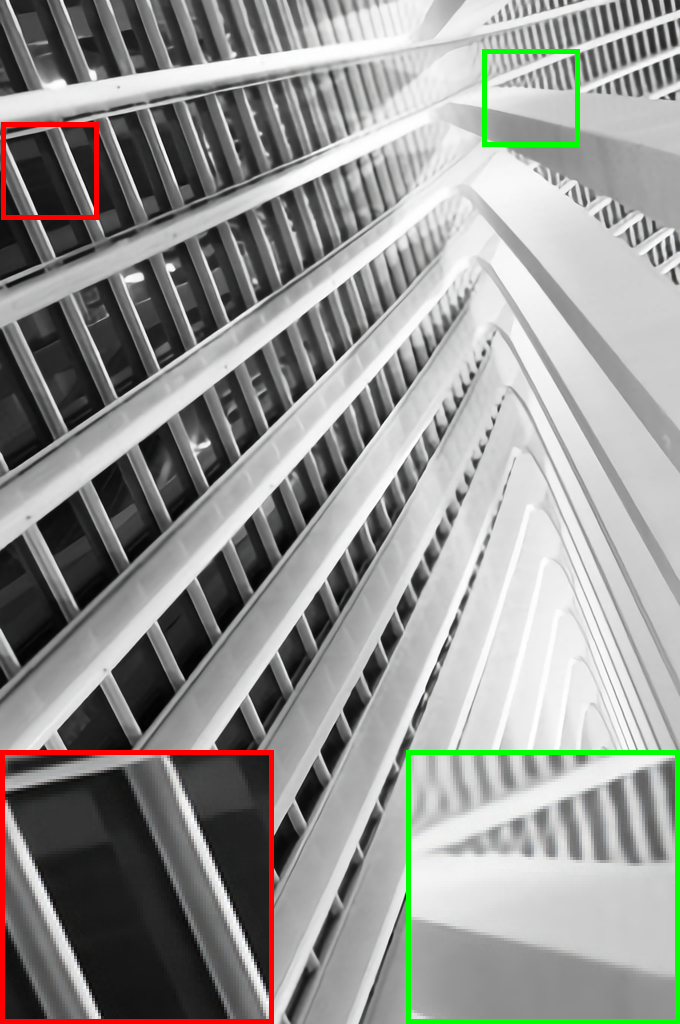} \\ 
    \subcaption*{\centerline{27.71/0.8784}} 
   \end{subfigure}
   \begin{subfigure}[H]{0.13\linewidth} 
    \subcaption*{\centerline{KL}} 
    \includegraphics[width=0.95\linewidth, height=1.3\linewidth]{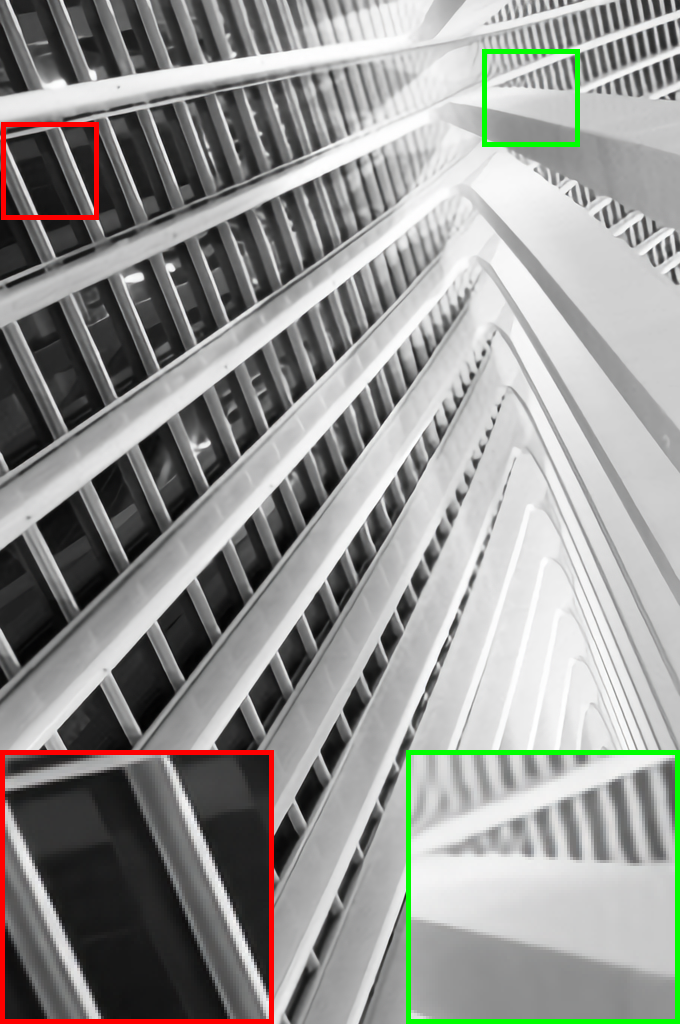} \\ 
    \subcaption*{\centerline{27.67/0.8786}} 
   \end{subfigure}
   \begin{subfigure}[H]{0.13\linewidth} 
    \subcaption*{\centerline{EQ}} 
    \includegraphics[width=0.95\linewidth, height=1.3\linewidth]{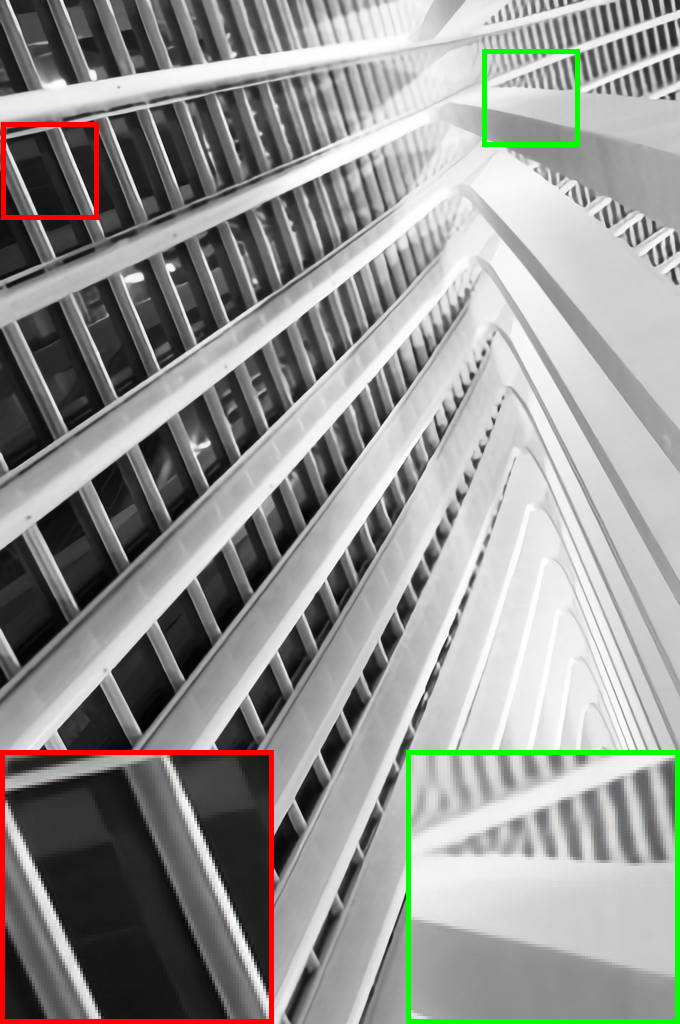} \\ 
    \subcaption*{\centerline{27.72/0.8797}} 
   \end{subfigure}
   \begin{subfigure}[H]{0.13\linewidth} 
    \subcaption*{\centerline{ADMM}} 
    \includegraphics[width=0.95\linewidth, height=1.3\linewidth]{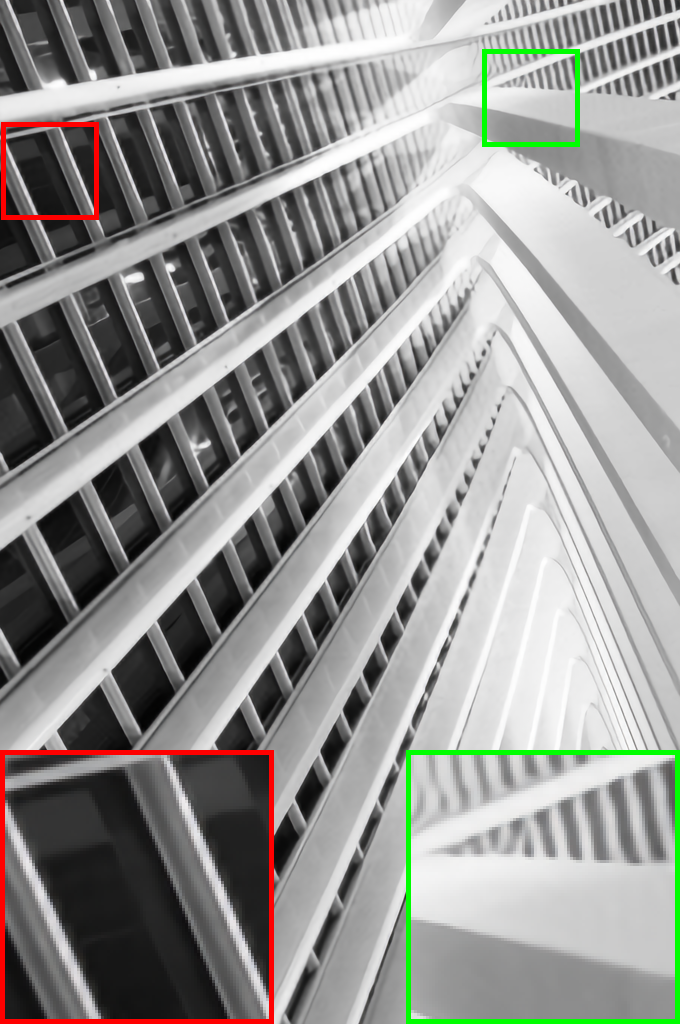} \\ 
    \subcaption*{\centerline{27.42/0.8756}} 
   \end{subfigure}
   \begin{subfigure}[H]{0.13\linewidth} 
    \subcaption*{\centerline{DQSS}} 
    \includegraphics[width=0.95\linewidth, height=1.3\linewidth]{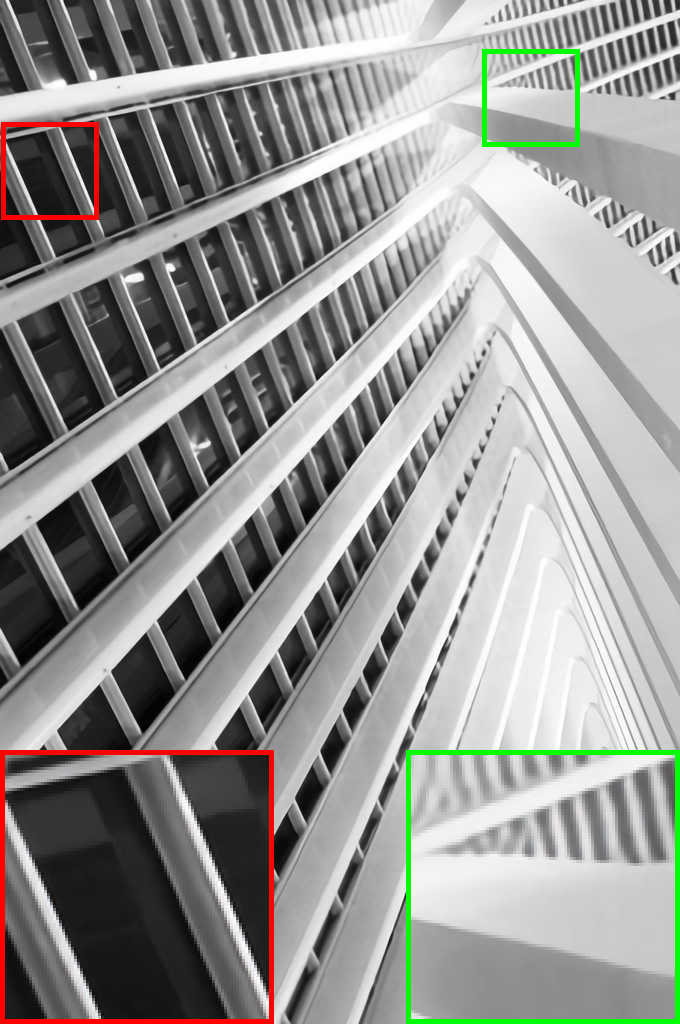} \\ 
    \subcaption*{\centerline{27.75/0.8803}} 
   \end{subfigure}
   \caption{Visual comparison of different methods on \textit{img042} from Urban100 for image super-resolution of 4x upscaling. MSRResNet is the backbone network.} 
   \label{fig:compare_msrresnet} 
\end{figure*} 

\begin{figure*}[!htb] 
   \centering
   \begin{subfigure}[H]{0.13\linewidth} 
    \subcaption*{\centerline{GT}} 
    \includegraphics[width=0.95\linewidth, height=0.75\linewidth]{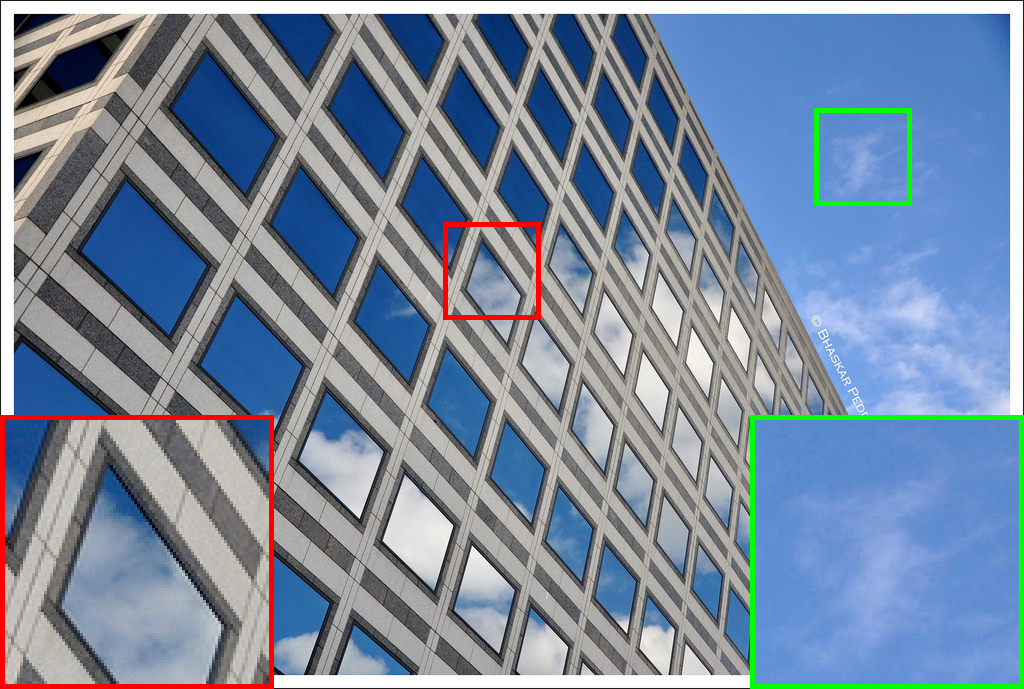} \\ 
    \subcaption*{\centerline{PSNR / SSIM}} 
   \end{subfigure}
    \begin{subfigure}[H]{0.13\linewidth} 
    \subcaption*{\centerline{FP32}} 
    \includegraphics[width=0.95\linewidth, height=0.75\linewidth]{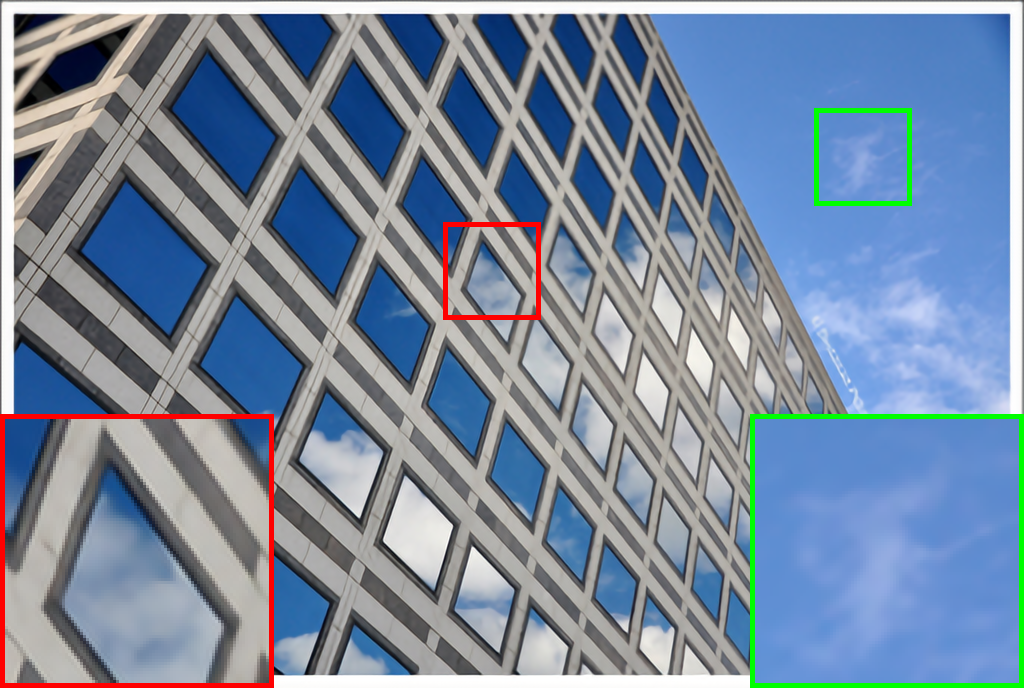} \\ 
    \subcaption*{\centerline{27.89/0.8593}} 
   \end{subfigure}
   \begin{subfigure}[H]{0.13\linewidth} 
    \subcaption*{\centerline{Max\_Abs}} 
    \includegraphics[width=0.95\linewidth, height=0.75\linewidth]{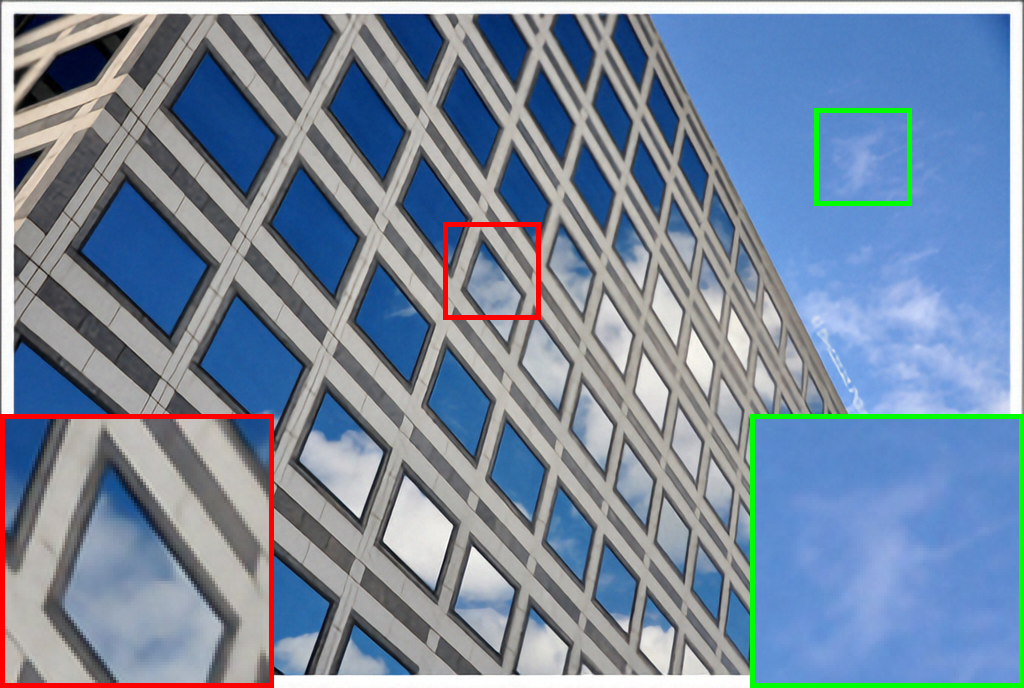} \\ 
    \subcaption*{\centerline{27.78/0.8508}} 
   \end{subfigure}
   \begin{subfigure}[H]{0.13\linewidth} 
    \subcaption*{\centerline{KL}} 
    \includegraphics[width=0.95\linewidth, height=0.75\linewidth]{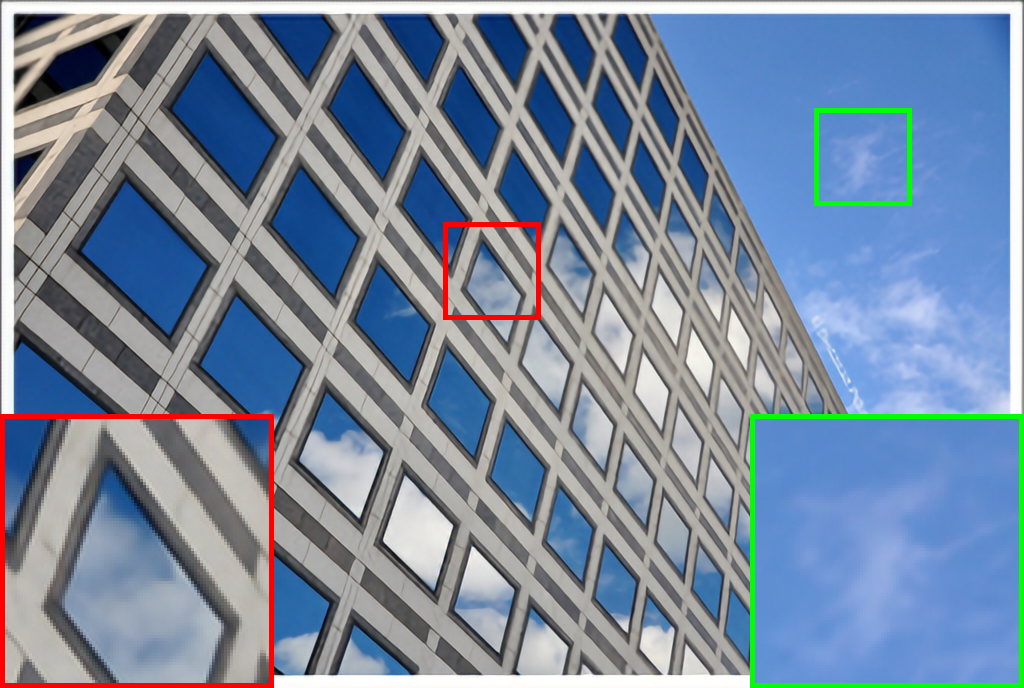} \\ 
    \subcaption*{\centerline{27.71/0.8557}} 
   \end{subfigure}
   \begin{subfigure}[H]{0.13\linewidth} 
    \subcaption*{\centerline{EQ}} 
    \includegraphics[width=0.95\linewidth, height=0.75\linewidth]{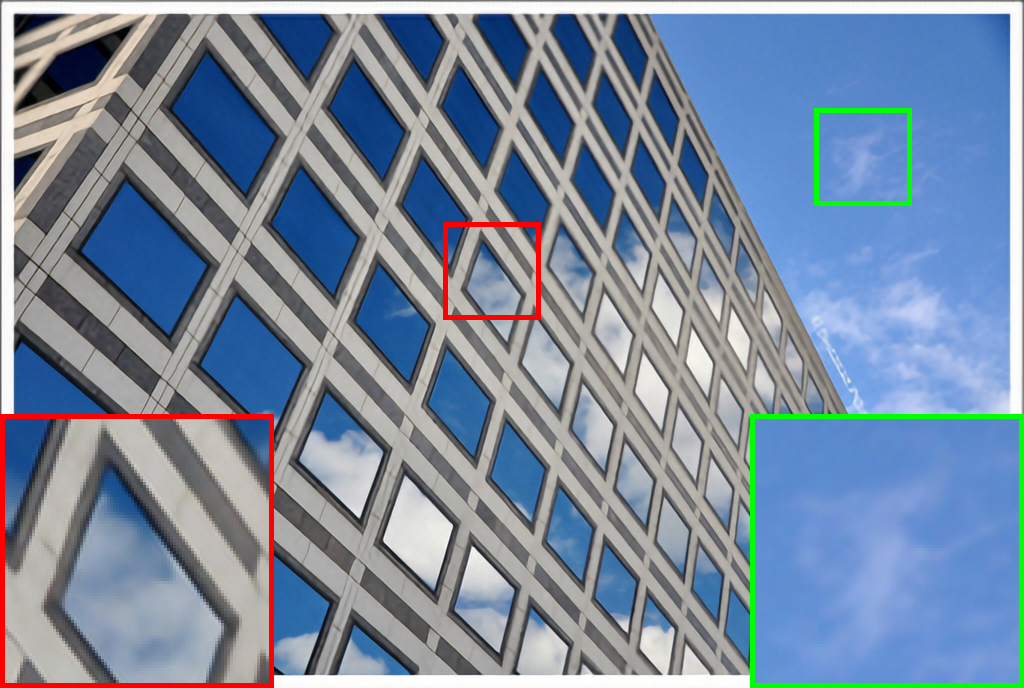} \\ 
    \subcaption*{\centerline{27.74/0.8539}} 
   \end{subfigure}
   \begin{subfigure}[H]{0.13\linewidth} 
    \subcaption*{\centerline{ADMM}} 
    \includegraphics[width=0.95\linewidth, height=0.75\linewidth]{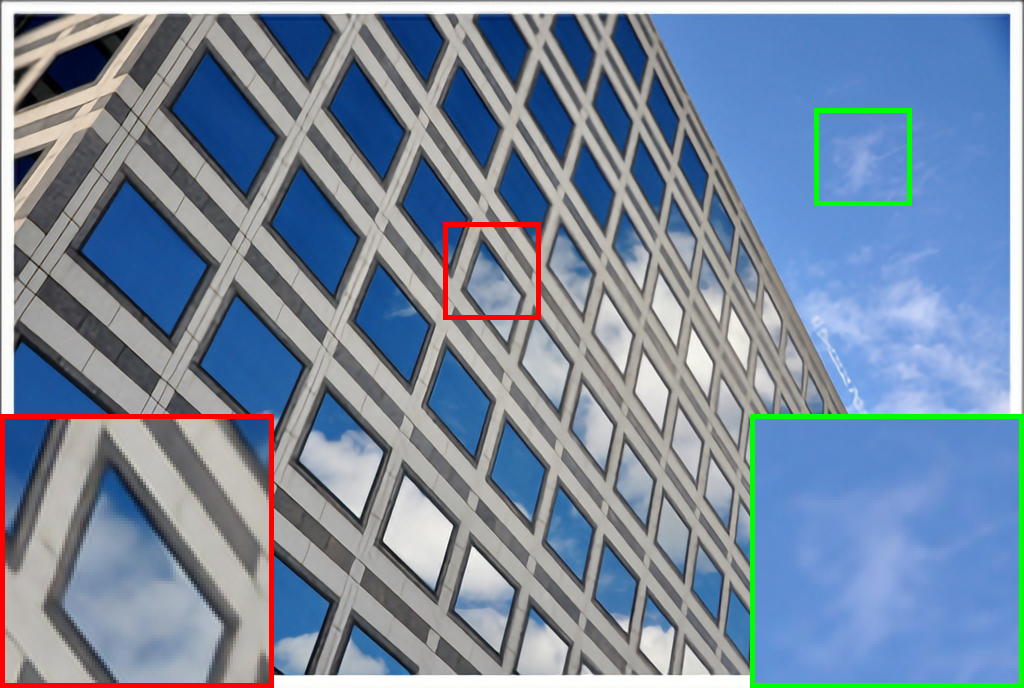} \\ 
    \subcaption*{\centerline{27.75/0.8554}} 
   \end{subfigure}
   \begin{subfigure}[H]{0.13\linewidth} 
    \subcaption*{\centerline{DQSS}} 
    \includegraphics[width=0.95\linewidth, height=0.75\linewidth]{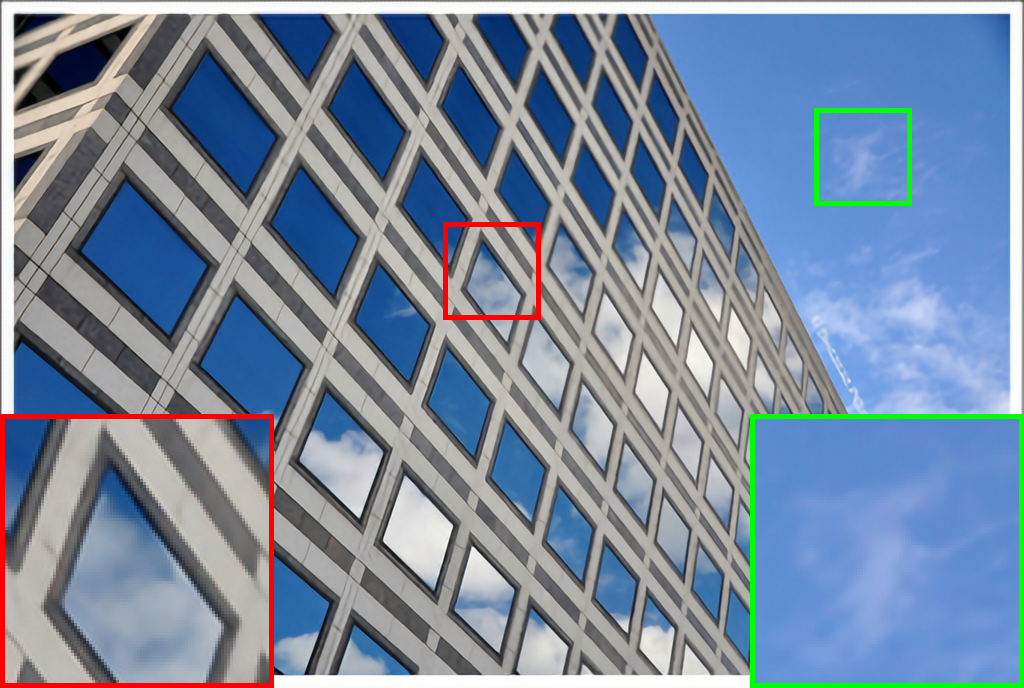} \\ 
    \subcaption*{\centerline{27.83/0.8562}} 
   \end{subfigure}
   \caption{Visual comparision of different methods on \textit{img035} from Urban100 for image super-resolution of 4x upscaling. EDSR is the backbone network.} 
   \label{fig:compare_edsr} 
\end{figure*}

\autoref{t1} presents the results of the 8-bit PTQ experiments on ImageNet classification. 
We can see from \autoref{t1} that DQSS always achieves the best accuracy. Under some circumstances, DQSS even outperforms FP32 models. We also notice that the performance gap between DQSS and other methods on VGG-16, ResNet-18 and ResNet-50 is not obvious. We assume that it is because the three networks are not sensitive to quantization. 
The assumption is verified in ShuffleNet-V2 and MobileNet-V2,  where the two networks are designed as lightweight networks and hard to be quantized. 
In these situations, all of the quantized methods cannot retain accuracy performance very well except DQSS. 
For example, DQSS has 71.628\% accuracy performance while Max\_Abs \cite{vanholder2016efficient} only has 63.440\% accuracy performance when the network refers to MobileNet-V2. For ADMM \cite{leng2018extremely}, which is the second best one, there still exists 1.215\% performance gap compared to DQSS.
We can also observe that the mean accuracy of DQSS across network architectures is as high as 71.577\% , which is very close to that in FP32 models, and much better than the others. This phenomenon reveals DQSS can not only quantize models with negligible performance degradation but also has good robustness across various network architectures.

\textbf{Experiments on image super-resolution:}
To demonstrate the superiority of the proposed method, we further apply DQSS to image super-resolution tasks. We use the DIV2K\cite{timofte2017ntire} for training, and five benchmark datasets including Set5 \cite{bevilacqua2012low-complexity}, Set14 \cite{ledig2017photo-realistic}, BSD100 \cite{martin2001a}, Urban100 \cite{huang2015single} and Manga109\cite{multimedia_aizawa_2020} for evaluation. The LR images are generated through bicubic downsampling. 
We evaluate the performance of DQSS by using the typical models for image super-resolution, \emph{i.e.,} MSRResNet\cite{wang2018esrgan} and EDSR\cite{Lim_2017_CVPR_Workshops}. 
Since compressing small models is more challenging, we choose the relatively smaller models in  BaiscSR\cite{wang2020basicsr} at the scale factor of  $\times4$  as the pretrained models
\footnote[1]{203\_EDSR\_Mx4\_f64b16\_DIV2K\_300k\_B16G1\_201pretrain\_wandb}
\footnote[2]{001\_MSRResNet\_x4\_f64b16\_DIV2K\_1000k\_B16G1\_wandb}. 

\autoref{tab:sr_MSRResNet_mx4} and \autoref{tab:sr_EDSR_mx4} show the quantitative results of the 8-bit PTQ experiments on image super-resolution. We can find that the proposed method almost outperforms the other quantized methods in all cases in terms of the objective evaluation metrics, i.e., PSNR and SSIM.
For example, when the experiments refer to MSRResNet on  Set5, the PSNR of DQSS is 0.23dB higher than that in ADMM \cite{leng2018extremely} and the SSIM gap between the two methods is 0.0023.
We can also see from  \autoref{tab:sr_MSRResNet_mx4} that the overall performance of ADMM \cite{leng2018extremely} is worse than that in Max\_Abs \cite{vanholder2016efficient}, which is not consistent with the results shown in \autoref{tab:sr_EDSR_mx4}. In fact, the different results also demonstrate the rationality of our motivation, i.e., existing quantization algorithms may outperform others in some networks, it is hard to judge that the algorithms are always better than others. 
\autoref{tab:sr_MSRResNet_mx4} and \autoref{tab:sr_EDSR_mx4} also show that the average quantitative metric across datasets of DQSS  is closest to that in FP32 models, which proves the robustness of DQSS in image super-resolution tasks.

Besides the quantitative comparison, we also present some images with rich texture and structure information, e.g., \textit{img035} and \textit{img042} from Urban100, to show the qualitative visual qualities. As shown in \autoref{fig:compare_msrresnet} and \autoref{fig:compare_edsr}, we can see that the visual results of DQSS can reconstruct the texture and structure information, and 
restore the edges successfully. For instance, the edges of the window of DQSS are sharper than other methods in \textit{img042}  as shown in \autoref{fig:compare_msrresnet}.
To sum up, DQSS can be applied to improving the quantized super-resolution networks as an effective method. 

\subsection{Experiments on QAT}
In this subsection, we conduct experiments on CIFAR-10 with 4-bit to validate the effectiveness of DQSS in QAT.
The state-of-the-arts including DoReFa\cite{zhou2016dorefa}, DSQ\cite{gong2019differentiable}, LSQ\cite{esser2019learned} and PACT\cite{choi2018pact} are selected as the candidate quantization strategies to set up the search space for DQSS. The initialized $\theta^{\alpha_i}$ and $\theta^{\beta_j}$ are also set to be 0.25 as there are also four methods.

\begin{table*}[htbp]
\small
\centering
\caption{\label{t4} The comparison of the Top-1 accuracy performance (\%) for the 4-bit QAT experiments on the CIFAR-10 classification. In each column, \textcolor{red}{red} / \textcolor{blue}{blue} represents \textcolor{red}{best} / \textcolor{blue}{second best}, respectively.}
\begin{tabular}{ccccccc}
\hline
\multirow {2} {*} {\diagbox{Methods}{Models}} & \multirow {2} {*} {VGG-16} & \multirow {2} {*} {ResNet-18} & \multirow {2} {*} {ResNet-50} & \multirow {2} {*} {ShuffleNet-V2} & \multirow {2} {*} {MobileNet-v2} & \multirow {2} {*} {Average architecture}  \\
\\
\hline
FP32 Baseline &94.17 &95.26 &95.47 &91.48 &92.32 &93.74\\
DoReFa\cite{zhou2016dorefa} & 93.84 & 94.81 & 95.31 & 89.61 & 90.87 &92.89  \\
DSQ\cite{gong2019differentiable}& 93.73 & 94.89 & {95.33} & 89.98 & 91.02 & 92.99 \\
LSQ\cite{esser2019learned} & 93.82 & 94.91 & \textcolor{blue}{95.41} & \textcolor{blue}{90.17} & \textcolor{blue}{91.22} & \textcolor{blue}{93.11}  \\
PACT\cite{choi2018pact} & \textcolor{blue}{94.02} & \textcolor{blue}{94.93} & 95.29 & 90.13 & 91.13 & 93.10 \\
DQSS & \textcolor{red}{94.33} & \textcolor{red}{95.27} & \textcolor{red}{95.62} & \textcolor{red}{90.89} & \textcolor{red}{91.88} & \textcolor{red}{93.60} \\
\hline
\end{tabular}
\end{table*}

We first train the networks which are utilized in \autoref{t1} on CIFAR-10 dataset to gain the FP32 baseline, and then compare DQSS with the SOTA methods. We adopt standard data augmentation techniques, namely random crop and horizontal flip. The SGD optimizer is used for both training phase and quantization fine-tuning phase. The learning rate $lr$ in training phase for weights is initialized as $10^{-2}$ while the learning rate $lr$ for weight parameters and the learning rate $lr_{\theta}$ for hyper-parameters in quantization fine-tuning phase are initialized as $10^{-3}$ and $10^{-4}$, respectively. 
We train a maximum 200 epochs for both training and fine-tuning processes, and decay for all learning rates by 10 at the beginning of the second hundred and the third hundred epochs. 
In addition, we apply a warm-up method for one epoch to observe the initialized quantization threshold.

\autoref{t4} presents the related results. We can find that the improvements brought by DQSS in QAT are not as obvious as those in \autoref{t1}. We suppose it is because the quantization fine-tuning process enhances the performance of other QAT methods.
Having said that, DQSS still outperforms the SOTA methods in all cases. For instance, our method outperforms DoReFa\cite{zhou2016dorefa} on VGG-16 by 0.49\%, and even outperforms the FP32 model by a slight 0.16\%.
Further, the advancement gets more apparent when the network refers to MobileNet-V2. In particular, DQSS surpasses DoReFa by 1.01\% on MobileNet-V2. 
Similar to \cite{li2021mqbench}, we also find that the performance of the state-of-the-arts is similar to each other, and none is significantly better than the others.
In the meanwhile, we consistently observe good performance of DQSS in terms of the mean accuracy over various network architectures, as shown in \autoref{t1}, \autoref{tab:sr_MSRResNet_mx4} and \autoref{tab:sr_EDSR_mx4}. This implies the DQSS still has the robustness in QAT methods.

\subsection{Ablation Study}
Besides the performance evaluation, we do some ablation studies to understand what enable DQSS to have the observed performance. We conduct ablation experiments on ImageNet classification by employing DQSS in PTQ methods because there is no fine-tuning process, and it is clear to identify the improvements brought by DQSS. 
Implementation details are kept the same as before.

We first evaluate the effect of the searching process, which is the essential of DQSS. We denote DQSS-None as the beginning state when the optimal mixed quantization strategies are not searched and each branch has the same importance  parameter, i.e., 0.25.
The related results are shown in \autoref{t5}. We observe that the searching process always leads to better performance. Taking into account the experiments are conducted on PTQ scenarios where the weights are not modified. The accuracy improvements can only gained from the searching process, which proves the searching process does work. Interestingly, we also find that DQSS-None performs slightly better than the second best algorithm in \autoref{t1}, i.e., ADMM \cite{leng2018extremely}. We suppose it is because the mean of multi quantized activations and multi quantized weights improve the expressiveness of the quantized network compared with the model quantized by a uniform quantization strategy.

\begin{table*}[htbp]
\small
\centering
\caption{\label{t5} The comparison of the Top-1 accuracy performance (\%) for ablation study.}
\begin{tabular}{ccccccc}
\hline
\multirow {2} {*} {\diagbox{Methods}{Models}} & \multirow {2} {*} {VGG-16} & \multirow {2} {*} {ResNet-18} & \multirow {2} {*} {ResNet-50} & \multirow {2} {*} {ShuffleNet-V2} & \multirow {2} {*} {MobileNet-v2} & \multirow {2} {*} {Average architecture}  \\
\\
\hline
DQSS & 71.842 & 70.294 & 75.679 & 68.442 & 71.628 & 71.577 \\
DQSS-None & 71.512 & 69.429 & 75.510 & 67.335 & 70.422 & 70.842 \\
\hline
\end{tabular}
\end{table*}

\begin{figure*}[!htb]
\centering
\includegraphics[width=160mm]{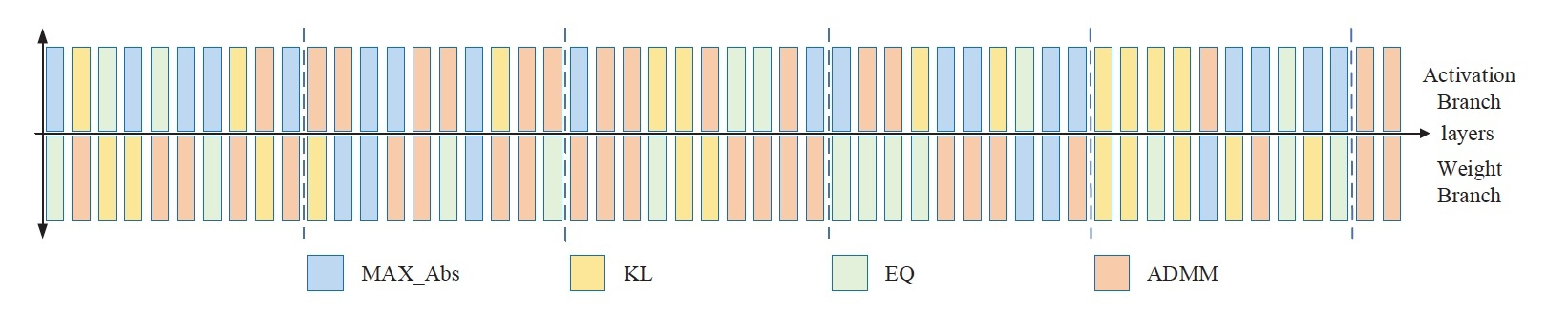}
\caption{The distribution of optimal quantization strategies for activations and weights on MobileNet-V2.}
\label{f4}
\end{figure*}

We also further explore what DQSS learns for the mixed quantization strategies assignment at last. Since the improvement between DQSS and DQSS-None in terms of MobileNet-V2 is the most apparent, we take the MobileNet-V2 as an example. \autoref{f4} presents the distribution of optimal quantization strategies for all layers. We  find that activations and weights can have different optimal quantization strategies. For example, Max\_Abs \cite{vanholder2016efficient} is selected for the activations while EQ \cite{wu2020easyquant} is selected for the weights in the first layer.
We can also see that ADMM \cite{leng2018extremely} is selected much more times than the other methods, followed by  KL\cite{vanholder2016efficient} and EQ \cite{wu2020easyquant}. On the contrary, Max\_Abs \cite{vanholder2016efficient} is selected as the optimal quantization strategy with the least times. The selected times is positively correlated with their performance in \autoref{t1} to some extent. In fact, the final assignment for mixed quantization strategies demonstrates that DQSS is able to select the higher-performance algorithms with more times as those algorithms lead to higher accuracy.

\section{Conclusion}
\label{section5}
In this paper, we propose a \textbf{d}ifferentiable \textbf{q}uantization  \textbf{s}trategy  \textbf{s}earch (DQSS), an automated framework for assigning optimal quantization strategy for individual layer, which does not require any domain experts and rule-based heuristics. 
DQSS can take advantages of the benefits of different quantization algorithms to keep the performance of quantized models as high as possible. Comprehensive experiments involving image classification and image super-resolution for both PTQ and QAT  show that DQSS could outperform the state-of-the-art quantization methods.

For future work, we would like to integrate more quantization algorithms into DQSS to enhance its effectiveness. 
To further validate the generality of DQSS,  evaluating on more tasks will be another future work. 

\bibliographystyle{unsrt}  
\bibliography{references}

\end{document}